\documentclass[letterpaper, 10 pt, conference]{ieeeconf}  % Comment this line out
\usepackage{graphics} % for pdf, bitmapped graphics files
\usepackage{epsfig} % for postscript graphics files
\usepackage{mathptmx} % assumes new font selection scheme installed
\usepackage{times} % assumes new font selection scheme installed
\usepackage{amsmath} % assumes amsmath package installed
\usepackage{amssymb}  % assumes amsmath package installed

\usepackage{amsthm}
\usepackage{bbm}
\usepackage{amsfonts,balance}
\usepackage{bm}
\usepackage{graphicx}
\usepackage{subfigure}
\usepackage{color}

\IEEEoverridecommandlockouts

\usepackage[svgnames]{xcolor} \definecolor{DarkGreen}{rgb}{0,0.5,0}
\definecolor{DarkRed}{rgb}{0.75,0,0}

\usepackage[pdftex,pdftitle={Transferable Pedestrian Motion Prediction Models at Intersections}]{hyperref}
\hypersetup{colorlinks,linkcolor={green!50!black},citecolor={green!50!black},urlcolor={blue!80!black}}
\makeatletter \let\NAT@parse\undefined \makeatother
\usepackage[sort,compress]{cite}

\title{\LARGE \bf
Transferable Pedestrian Motion Prediction Models at Intersections}%  \todo{general comment: you need to work on the wording, both in the way you compare with other algorithms, and in fixing grammatical errors}

\author{Macheng Shen$^{1}$, Golnaz Habibi$^{2}$, and Jonathan P.\ How$^{2}$% <-this % stops a space
\thanks{$^{1}$M. Shen is with the Department of Mechanical Engineering, Aerospace Controls Laboratory (ACL),
        Massachusetts Institute of Technology (MIT), 77 Massachusetts Ave., Cambridge, MA, USA.
        {\tt\small macshen@mit.edu}}%
\thanks{$^{2}$G. Habibi and J. P. How are with the Department of Aeronautics and Astronautics, ACL, MIT
        {\tt\small \{golnaz,jhow\}@mit.edu}%
}}

\begin{document}

\maketitle
\thispagestyle{empty}
\pagestyle{empty}

%%%%%%%%%%%%%%%%%%%%%%%%%%%%%%%%%%%%%%%%%%%%%%%%%%%%%%%%%%%%%%%%%%%%%%%%%%%%%%%%

%%%%%%%%%%%%%%%%%%%%%%%%%%%%%%%%%%%%%%%%%%%%%%%%%%%%%%%%%%%%%%%%%%%%%%%%%%%%%
\begin{abstract}
One desirable capability of autonomous cars is to accurately predict the pedestrian motion near intersections for safe and efficient trajectory planning. We are interested in developing transfer learning algorithms that can be trained on the pedestrian trajectories collected at one intersection and yet still provide accurate predictions of the trajectories at another, previously unseen intersection. We first discussed the feature selection for transferable pedestrian motion models in general. Following this discussion, we developed one transferable pedestrian motion prediction algorithm based on Inverse Reinforcement Learning (IRL) that infers pedestrian intentions and predicts future trajectories based on observed trajectory. %For evaluation, we first compared the accuracy of the proposed algorithm with that of augmented semi-nonnegative sparse coding (ASNSC) and context-based ASNSC (CASNSC), trained and tested at the same intersection. We then compared the accuracy of the proposed algorithm tested at another unseen intersection with that of ASNSC both trained and tested at this unseen intersection, as a baseline. The result shows that the proposed algorithm achieves comparable accuracy with ASNSC and CASNSC and has good transferring capability. 
We evaluated our algorithm at three intersections. We used the accuracy of augmented semi-nonnegative sparse coding (ASNSC), trained and tested at the same intersection as a baseline. The result shows that the proposed algorithm improves the baseline accuracy by a statistically significant percentage in both non-transfer task and transfer task.
\end{abstract}
\section{Introduction}
%\todo{General comment: you have used IRL , so at least you have to compare and discuss what is your contribution in this field}
%\todo{do not copy from abstract, also the introduction should be much longer}
One desirable capability of autonomous cars is accurately predicting the motion of nearby pedestrians to ensure safe and efficient trajectory planning. Learning based methods that discover the patterns of pedestrian trajectories have demonstrated success in trajectory prediction \cite{chen2016augmented,kitani2012activity,joseph2011bayesian}. Nonetheless, in order to predict human motion at a specific scene, most of these approaches require motion data collected at this scene to train a predictive model, which requires a significant amount of hardware investment and effort. Moreover, it is undesirable and sometimes infeasible to store all the model parameters for every intersection. A more useful approach is to learn a model that is \textit{transferable} from one scene to another, possibly with different geometric scales and topologies. One significant benefit of this transfer learning approach is that it saves a lot of expense on data collection. 
%Moreover, this type of approach can be trained on data collected at different scenes, which could potentially result in %higher prediction accuracy than that of the non-transferable approaches trained at a single scene.

\indent We are interested in developing transferable pedestrian motion prediction algorithms at intersections that achieve reasonable prediction accuracy at unseen intersections. The difficulty of transferring is that trajectories collected at different intersections could exhibit quite different patterns, depending on the geometry and semantics at these intersections. A transferable prediction algorithm is expected to discover the relationship between the trajectories and the geometric and semantic properties of the intersections, instead of focusing on the trajectories alone. 
\begin{figure}[t!]
\centering
\includegraphics[width=1\columnwidth]{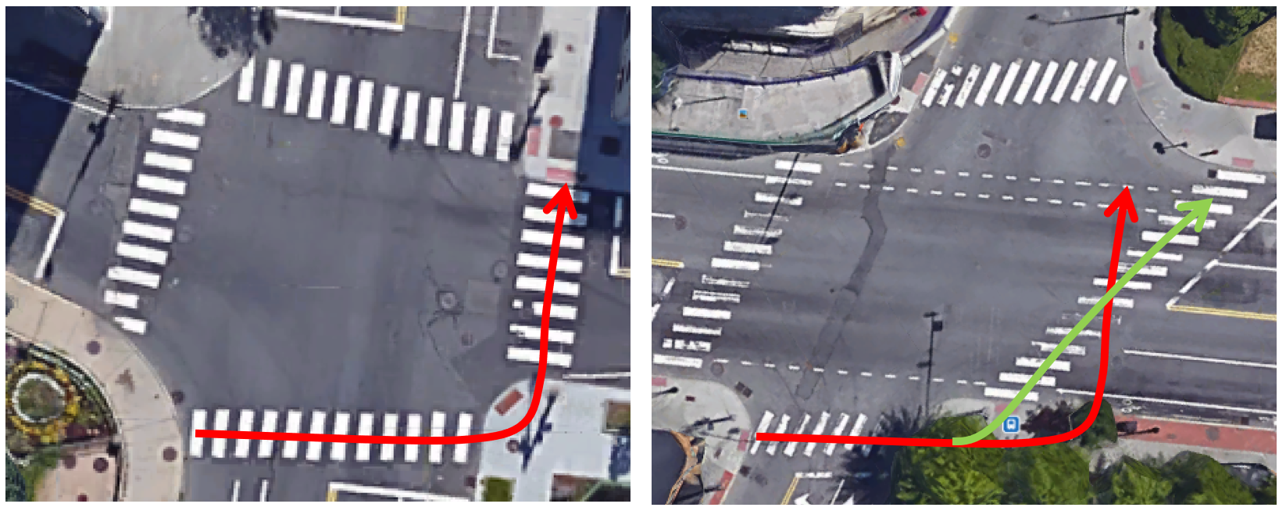}  
\caption{Example of transferring pedestrian motion model: The left figure shows the training intersection; The right figure shows the test intersection, where the green arrow represents a reasonable prediction, while the red arrow is an unacceptable prediction that simply copies the trajectory from the training intersection. Image from Google Earth.}
\label{key_issue}
\end{figure}

\indent Several existing works addressed human motion prediction, for example, \cite{ballan2016knowledge,chen2016augmented,karasev2016intent,kitani2012activity,fouhey2014predicting,joseph2011bayesian}. We divide the previous works based on the representation of the pedestrian motion into the following three main categories: 
\begin{enumerate}
\item The first is velocity-based model: in velocity based models \cite{chen2016augmented,joseph2011bayesian}, the pedestrian motion pattern is represented as a parametrized 2-dimensional velocity field, where the parameters are learned from trajectories. One crucial assumption in most of this type of models is that the velocity field can be decomposed into two uncorrelated velocity components along the two orthogonal directions defined by the coordinate frame. A coordinate frame (sometimes along with a discretization of the domain as in \cite{chen2016augmented}) must be selected before parameter learning.

\item The second is state-based model: in state-based models \cite{vasquez2009incremental,makris2002spatial,chen2016augmented}, the pedestrian motion pattern is represented as transition between states, which are either learned or prescribed based on the semantics. The states are very scene-specific, and sometimes the number of states could be large. The transition between the states is typically modeled with a Hidden Markov Model (HMM), and the transition matrix is learned from the data.
\item  The third is Markov Decision Process (MDP)-based model: in MDP based models \cite{bruce2004better,karasev2016intent,kitani2012activity,ziebart2009planning}, the pedestrian motion is modeled as a MDP, and the pedestrian motion is induced by the reward function, which is typically learned through inverse reinforcement learning (IRL) \cite{ramachandran2007bayesian} from the trajectories or prescribed based on the semantics.
\end{enumerate}

Intuitively, models of the first two categories are not well-suited for transfer. For example, as illustrated in Fig.~\ref{key_issue}, a velocity based model would predict a trajectory similar to that at the training intersection (red arrow), regardless of the geometry at the test intersection, which is not a reasonable prediction. As for the state-based model, mapping the states at the training intersection to the test intersection could be difficult. In contrast, models of the third category would predict a trajectory that follows the semantic context at the test intersection (green arrow), which is a reasonable prediction.

While the MDP based methods could be transferable, the authors are not aware of previous work that has discussed the design rule of a transferable prediction model in general. We believe such a discussion has a significant practical value for model selection, which will be presented in this paper, together with an inverse reinforcement learning (IRL) based method.

The main contribution of this paper is that we propose a transferable model based on IRL that can achieve high accuracy in both transfer and non-transfer tasks. In Section II, we discuss the feature selection for transferable motion prediction, based on which we conclude that MDP based model is a suitable framework for transfer. We also discuss a problem of the previous MDP based models that lead to low prediction accuracy. In Section III, we present an MDP based model and Expectation-Maximization (EM) based learning algorithm that overcomes the aforementioned problem. In Section IV, we show prediction results of the proposed algorithm and compare the prediction accuracy with that of augmented semi-nonnegative sparse coding (ASNSC). In Section V, we summarize the results and contributions of this paper.

%\includegraphics[width=\textwidth]{figure/tra11.png}   
%\caption{Illustration of the over-confidence issue in constant reward models}
%\label{tra11}
 
\section{Feature selection for transferable model}

A pedestrian motion model can be abstracted as a mapping from a feature vector to an output, $\mathcal{M}(\theta): \psi(x)\rightarrow Y$, where we consider a parametrized model with parameter $\theta$, and $x$ is the state of the pedestrian. In many cases, this model has a specific representation corresponding to the specific coordinate frame $\mathcal{F}$ that we use to describe the pedestrian motion. For example, in the Dirichlet Process Gaussian Process (DPGP) \cite{joseph2011bayesian} model, the feature vector is the $x$-$y$ position of the pedestrian in the frame $\mathcal{F}$. The output is the $x$-$y$ velocity components in the frame $\mathcal{F}$ at that position. The model parameters $\theta_{\mathcal{F}}$ include the GP hyper-parameters, the DP mixture weights, and the mobility pattern assignments.

In general, these model parameters, as well as the feature and output representation depend on the chosen coordinate frame $\mathcal{F}$. Suppose we represent the same motion model using a different coordinate frame $\mathcal{F}'$ that only differs from $\mathcal{F}$ with a rigid body transformation, the model parameters have to be transformed accordingly in order to keep the model unchanged.

Nevertheless, for some specific model and coordinate transformation, it could be impossible to preserve the model. For example, in the DPGP model, if we apply a rotation to the frame $\mathcal{F}$ to get the transformed frame $\mathcal{F}'$, retaining the same model is impossible as GP assumes uncorrelated $x$-$y$ velocity components. %We prove this proposition using proof by contradiction:
%\begin{prop}
%A motion model represented by $GP(\theta)$ defined with a coordinate frame $\mathcal{F}$ cannot always be represented by $GP(\theta')$ defined with a rotated coordinate frame $\mathcal{F}'$, for an arbitrary rotation angle $\Delta \theta=\theta-\theta'$.
%\end{prop}
%\begin{proof}
%Suppose there exists a $GP(\theta')$ in the frame $\mathcal{F}'$ that corresponds to the same motion model represented by $GP(\theta)$ in the frame $\mathcal{F}$, The following relationship exists between the velocity components corresponding to these two representation:\\
%\begin{equation}
%\begin{aligned}
%&v_{x'}=v_xcos(\Delta \theta)-v_ysin(\Delta \theta),\\
%&v_{y'}=v_xsin(\Delta \theta)+v_ycos(\Delta \theta).
%\end{aligned}
%\end{equation}
%\indent As a result of the uncorrelation assumption, 
%\begin{equation}
%\mathbb{E}[v_{x'}v_{y'}]=(\mathbb{E}[v_x^2]-\mathbb{E}[v_y^2])sin(\Delta \theta)cos(\Delta \theta)=0.
%\end{equation}
%\indent In order for this to hold for any arbitrary rotation angle, the following has to hold:
%\begin{equation}
%\mathbb{E}[v_x^2]=\mathbb{E}[v_y^2].
%\label{contra}
%\end{equation}
%\indent Eq. (\ref{contra}) is a strong constraint on the velocity field represented by $GP(\theta)$, and it does not hold for an arbitrary velocity field. Therefore, it is not guaranteed that a GP motion model can be preserved under a rotation transformation of the coordinate frame. 
%\end{proof}

The fact that model parameters typically depend on the selected coordinate frame does not cause any substantial difficulty for motion prediction in the same intersection. However, it is an undesirable property for transferring the learned model to an unseen intersection. In general cases, a global coordinate transformation between the two intersections does not always exist, especially when the topologies of the two intersections are different. Therefore, we impose the following requirement for model selection: The feature vector, the model parameter, and the output representation are invariant under a rigid body transformation of the coordinate frame. In order to satisfy this requirement, it suffices if both the feature vector and the output representation are invariant under the transformation. 
Thus we propose the following: \\
\noindent {\bf Remark:} If the feature vector $\psi(x)$ and the output $Y$ of a motion model are invariant under any arbitrary coordinate transformation, then the model parameter $\theta$ is also invariant under any arbitrary coordinate transformation.

We justify this statement with the following reasoning. Suppose the model parameter corresponding to the new coordinate frame $\theta'$ is different from the original parameter $\theta$, then there exists an $x$ such that the output $Y'$ corresponding to the input $\psi'(x)$, under the mapping $\mathcal{M(\theta')}$ is different from the output $Y$ corresponding to the input $\psi(x)$ under the mapping $\mathcal{M(\theta)}$. However, this violates the pre-condition that the output $Y$ is invariant under the coordinate transformation. Therefore, $\theta'=\theta$ must hold.

For pedestrian motion model, some typical features that are invariant under coordinate transformation are: semantic context label, reward function, angle between two curbsides (this can be seen as the inner product of two unit vectors along the curbsides, which is invariant), distance from a point to a curbside (this can be seen as the norm of a vector, which is invariant).

Based on the above discussion, we conclude that MDP model and IRL is a suitable framework for transferable motion model learning, as the pedestrian motion is summarized by the reward function, which is scalar function invariant to coordinate transformation.\\
\section{Transferable MDP pedestrian motion model}
\subsection{Preliminaries}
We model the pedestrian trajectory as a result of planning sub-optimally according to an MDP, which is defined as a tuple of state $S$, action $A$ and reward $R$. We consider the following variables into the model $S: (x,v,\lambda),$ where $x$ is a vector representing the 2-D position of the pedestrian; $v$ is a scalar representing the current velocity magnitude, which we assume is observed; and $\lambda$ is the latent intent of the pedestrian. The action is the heading angle of the pedestrian 
$A: \phi.
$
Reinforcement learning \cite{sutton1998reinforcement} is a technique to solve the MDP. Given the reward function $R$ and an optimal policy $\pi^*$ that maps each state $x$ to an optimal action $a$, the optimal value function $V(x)$ and the cooresponding Q-function are defined recursively as
\begin{equation}
\begin{aligned}
&V^{\pi^*}=max_u Q^{\pi^*}(x,u),\\ Q^{\pi^*}&(x,u)=R(x,u)+\gamma V^{\pi^*}(x+\pi^*).
\end{aligned}
\end{equation}
The task of reinforcement learning is to find an optimal policy given the reward function, while the task of inverse reinforcement learning is to infer the reward function from the demonstrated trajectories \cite{abbeel2011inverse}.

\begin{figure}[t]
  \centering
  \includegraphics[width=0.8\columnwidth]{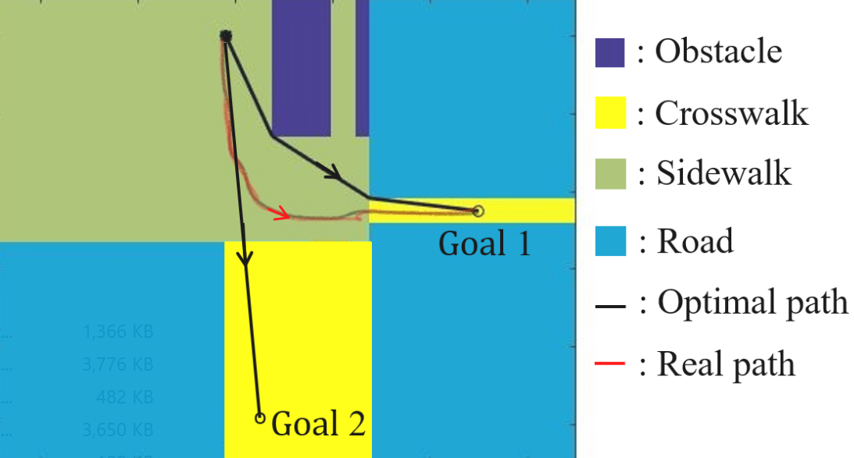}   
  \caption{Illustration of the over-confidence issue in the piece-wise constant reward models: The two black lines are the optimal trajectories under the piece-wiese constant reward function model, while the red line represents a real trajectory}
  \vspace{-0.3cm}
  \label{illustration}
\end{figure}

In the previous works with MDP models \cite{kitani2012activity,karasev2016intent}, designed for long term prediction, is not well suited for predictions at medium scale scenes such as intersections, where a more refined parametrization is expected to have a better accuracy. In the previous formulations, the reward function field is piece-wise constant that only varies across the boundary between two different semantic contexts. A consequence of this model is that the optimal trajectories are piece-wise straight lines, which change directions only upon the boundaries between two different semantic contexts. Nonetheless, real pedestrian trajectories seldom look like piece-wise straight lines. As a result, the prediction accuracy of the trajectory would be low. Moreover, the prediction accuracy of the high-level intention would also be low, which can be illustrated with the example shown in Fig. \ref{illustration}. Suppose the pedestrian has two potential goal locations as shown in Fig. \ref{illustration}, and the actual intention of the pedestrian is Goal 1. \\
\indent In order to infer the intention of the pedestrian, we use Bayesian rule 
\begin{equation}
p(\lambda|t)\propto p(t|\lambda)p(\lambda),
\end{equation} 
where $\lambda=1,2$ is a discrete variable describing the true intention of the pedestrian, and $t$ denotes the pedestrian trajectory.\\
\indent In the beginning, the real path aligns well with the optimal path corresponding to Goal 2. As a result,
\begin{equation}
p(t|\lambda=2)\gg p(t|\lambda=1).
\end{equation} 
Since we have no prior knowledge of the true intention of the pedestrian before we observe the partial trajectory, we assign a uniform prior
\begin{equation}
p(\lambda=1)=p(\lambda=2)=0.5
\end{equation}
Combining these two facts, we obtain 
\begin{equation}
p(\lambda=2|t)\gg p(\lambda=1|t),
\end{equation} which is an incorrect over-confident inference.

In this work, we made two major modifications to fix this problem: (1) The reward function at one specific location not only depends on the local semantic context  but also on the surrounding semantic contexts. (2) Abrupt turning of heading angle is penalized. In the following sections, we show the detailed formulation of the proposed IRL model.

\subsection{Reward Parametrization and Feature Construction}
\begin{figure}[t]
\centering
\includegraphics[width=0.8\columnwidth]{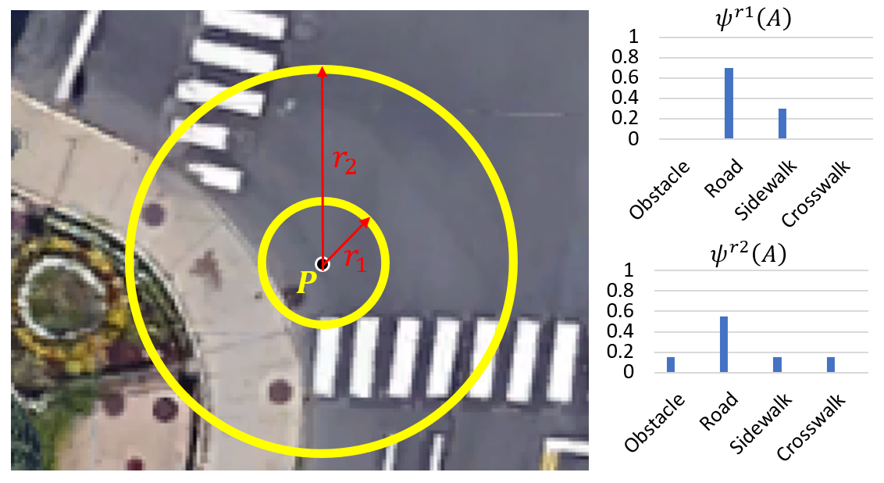}   
\caption{Illustration of the feature construction: the histogram feature of the semantic contexts are extracted at the concentric circles with radius $r1$ and $r2$ around point P}
\label{sem_illu}
%\includegraphics[width=\textwidth]{figure/tra11.png}   
%\caption{Illustration of the over-confidence issue in constant reward models}
%\label{tra11}
  \vspace{-0.3cm}
\end{figure}

We decompose the reward function into two components
\begin{equation}
R(x;\lambda)=R_x(x)+R_{\lambda}(x),%+R_{\dot{\phi}}(\dot{\phi}),
\end{equation}
where $R_x(x)$ is only dependent on the local and neighboring semantic contexts. For example, the reward on a road context would be lower than that on a crosswalk context; $R_{\lambda}(x)$ is dependent on the distance from the current position to the goal location. Positions that are closer to the goal location will have higher rewards.\\
\indent We observed, however, that pedestrians do not always prefer walking on sidewalks or crosswalks than on roads. For example, some pedestrians frequently walk on the roads near sidewalks and crosswalks, but none of them tend to walk on roads far away. In order to capture this behavior, we parametrize $R_x(x)$ as follows.
\begin{itemize}
\item  First, we classify the intersection into four different semantics: (1) obstacles (e.g., buildings, bushes, etc.), (2) road, (3) sidewalk, (4) crosswalk.
\item Then we record the semantic label of point $x$ as well as the histogram of the semantics on two concentric shells regions centered at $x$ with radius $r_1$ and $r_2$, as proposed in \cite{ballan2016knowledge}. In this work, we use $r_1=1\mbox{ }m$ and $r_2=3\mbox{ }m$ (see Fig. \ref{sem_illu}).
\end{itemize}

We then construct the feature vector $\psi(x)$ described as follows:
\begin{equation}
\psi(x)=(\psi_1(x)^T,\psi_2(x)^T,\psi_3(x)^T)^T,
\end{equation}
where $\psi_1(x)$ is the one-hot encoding of the semantic label at point $x$, which is a 4 by 1 vector; $\psi_2(x)$ and $\psi_3(x)$ are conditional histogram feature defined as follow:
\begin{equation}
\psi_2(x)=
 \begin{cases}%{rcl}
 (\psi^{r1}(x)^T,\psi^{r2}(x)^T)^T, &\psi_1(x)=(0,1,0,0)^T\\
 \bm{0}_{8\times 1}, &otherwise
 \end{cases},
\end{equation}

\begin{equation}
\psi_3(x)=
 \begin{cases}
 (\psi^{r1}(x)^T,\psi^{r2}(x)^T)^T, & \psi_1(x)=(0,0,1,0)^T\\
  & or\quad \psi_1(x)=(0,0,0,1)^T\\
  \bm{0}_{8\times 1}, &otherwise
 \end{cases}
\end{equation}
where $\psi^{r1}(x)$ and $\psi^{r2}(x)$ are the normalized histogram vector at point $x$ with radius $r_1$ and $r_2$, respectively.

We illustrate the feature construction with an example (point P) shown in Fig. \ref{sem_illu}: As the point $A$ is on the road, $\psi_1(A)^T=(0,1,0,0)$. As a result, $\psi_2(A)=(\psi^{r1}(x)^T,\psi^{r2}(x)^T)^T$ and $\psi_3(A)=\bm{0}_{8\times 1}$. The feature vector $\psi(x)_{20\times 1}$ is a function of the semantic context at that point, which is independent of the coordinate frame, thus satisfying the requirement proposed in Section~II.
We parametrize $R_x(x)$ as a linear function of $\psi(x)$
\begin{equation}
R_x(x)=w^T\psi(x),
\end{equation}
where $w$ is a sparse vector with only several non-zero entries, which are $w_1, w_2, w_7, w_8, w_{11}, w_{12}, w_{13}, w_{14}, w_{17}, w_{18}$. The other entries that are enforced to be zero all have concrete interpretations. For example, $w_3=0$ means that walking on sidewalk is not penalized.

 We model $R_{\lambda}(x)$ with a radial function:
\begin{equation}
R_{\lambda}(x)=K(x-x_{\lambda}),
\end{equation}
where $K(x-x_{\lambda})$ is a radial function centered at $x_{\lambda}$. For example, here we use
\begin{equation}
K(x-x_{\lambda})=\left\{
 \begin{array}{rcl}
 -\frac{1}{2}\sqrt{||x-x_{\lambda}||-d},&      & ||x-x_{\lambda}||>d\\
  0,&      &||x-x_{\lambda}||\leq d
 \end{array} \right.,
\end{equation}
where $x_{\lambda}$ is the goal location, and $d\geq 0$ is the finite radius of the goal. Both $x_{\lambda}$ and $d$ should be determined based on the semantic context around the intersection. $R_{\lambda}(x)$ also does not depend on the coordinate, as it is a function of a scalar quantity $||x-x_{\lambda}||$.
\subsection{EM Parameter Learning}
We want to learn the maximum likelihood estimator for the model parameters that maximizes the data likelihood $p(t|R)$, where $t$ represents the trajectories. We adopt the modeling approach typically used in IRL, i.e., assuming that the agent is behaving sub-optimally  and the probability of a sub-optimal behavior is proportional to the exponential of the corresponding Q-function. Moreover, we also add an additional term $\tilde{R}=-C_{\phi}\tanh(\beta|\Delta\phi_k|^{\alpha})$ to the Q-function to penalize turning of heading angle
\begin{equation}
L(R;t)=p(t|R)=\sum_{\lambda}p(t|R,\lambda)p(\lambda|R),
\end{equation}
where
\begin{equation}
\log(p(t|R,\lambda))=\sum^{N_t}_{i=1}\sum^{N_i}_{k=1}\log\left[\frac{\exp(\eta \tilde{Q}(x_k,\phi_k;\lambda_i))}{\sum^{N_a}_{j=1}\exp(\eta \tilde{Q}(x_k,a_j;\lambda_i))}\right],
\label{log-ll}
\end{equation}
$\tilde{Q}(x_k,\phi_k;\lambda_i)=Q(x_k,\phi_k;\lambda_i)-C_{\phi}\tanh(\beta|\Delta\phi_k|^{\alpha})$ is the surrogate Q-function that penalizes turning of heading angle, and $Q(x_k,\phi_k;\lambda_i)$ is the original Q-function that is obtained by solving the MDP defined in section III.C; $N_t$ is the total number of trajectories; $N_i$ is the data length of the $i$th trajectory; $N_a$ is the number of discretization for the heading angle; $\eta$ is the proficiency level of the demonstrator.
To evaluate Eq.~\ref{log-ll}, we need to know the latent intention variable $\lambda$ for each trajectory, which itself should be inferred from the trajectory. Therefore, we formulate an EM learning algorithm.
\subsubsection{EM for parameter learning}
In order to maximize the log-likelihood Eq. (\ref{log-ll}), we also need the posterior over the intent of each trajectory. We model it as follows:
\begin{equation}
\begin{aligned}
&p(\lambda_i|R)=\text{softmax}\left(\frac{\sum^{N_i}_{k=1}\log(Q_a(x_k,\phi_k;\lambda_i))}{\sqrt{N_i}+10}\right),\\
&Q_a(x_k,\phi_k;\lambda_k)=\frac{\exp(\eta \tilde{Q}(x_k,\phi_k;\lambda_i))}{\sum^{N_a}_{j=1}\exp(\eta \tilde{Q}(x_k,a_j;\lambda_i))}.
\end{aligned}
\label{intent_post}
\end{equation}
Instead of maximizing Eq. (\ref{log-ll}), which is not well defined without given $\lambda$, we maximize the expected log-likelihood over the reward function parameter $\theta$:
\begin{equation}
\underset{\theta}{\text{maximize}} \sum_{\lambda}\log(p(t|R_{\theta},\lambda))p(\lambda|R_{\theta}).
\end{equation}

\subsubsection{Constraints for model parameters}
The model parameters hat must be learned from the data are $\theta=\{w,C_{\phi},\beta,\alpha,\eta\}$. All of these parameters have concrete interpretations, and, therefore, we must specify some constraints on the model values to ensure the result is reasonable. We specify the following constraints and explain their interpretations below:
\begin{equation}
\begin{aligned}
&(1)  w_1=-2.5,\quad (2) w_2\leq-0.5, \quad(3) w_7=w_8\geq 0,\\
&(4) w_{11}=w_{12} \geq 0, \quad(5)w_{13}\leq 0, \quad(6) w_{14}\leq 0,\\
&(7) w_{17}\leq 0, \quad(8) w_{18}\leq 0,\quad (9) 2w_2+w_7+w_{11}\leq w_{14}+w_{18},\\
&(10) C_{\phi}\geq 0, \quad (11) \beta \geq 0,\quad (12) \alpha \geq 0, \quad(13)\eta \geq 0.
%&(1) \quad w_1=-2.5\\
%&(2)\quad w_2\leq-0.5\\
%&(3)\quad w_7=w_8\geq 0\\
%&(4)\quad w_{11}=w_{12} \geq 0\\
%&(5)\quad w_{13}\leq 0\\
%&(6)\quad w_{14}\leq 0\\
%&(7)\quad w_{17}\leq 0\\
%&(8)\quad w_{18}\leq 0\\
%&(9) \quad 2w_2+w_7+w_{11}\leq w_{14}+w_{18}\\
%&(10)\quad C_{\phi}\geq 0\\
%&(11)\quad \beta \geq 0\\
%&(12)\quad \alpha \geq 0\\
%&(13)\quad \eta \geq 0
\end{aligned} \label{eq-const}
\end{equation}
Constraint (\ref{eq-const}-1) specifies the reward for an obstacle context. This is set as a low value to prevent the modeled trajectory entering an obstacle. (\ref{eq-const}-2) specifies a constraint for the reward of a road context. This constraint is required in case that the training trajectories are not rich enough to learn to assign a low reward to the road context. On the training data we used in this work, we found that the strict inequality $w_2<-0.5$ holds for the optimizer, so constraint (\ref{eq-const}-2) actually does not affect the learning result. Constraints (\ref{eq-const}-3) and (\ref{eq-const}-4) ensure that a road context that is near a crosswalk or sidewalk has a higher reward than the ones that are not. (\ref{eq-const}-5)--(\ref{eq-const}-8) ensure that a crosswalk or sidewalk context that is near an obstacle or road has a lower reward than ones that are not. (\ref{eq-const}-9) ensures that a road context right beside a crosswalk or sidewalk has a lower reward than a crosswalk or sidewalk context right beside a road.  (\ref{eq-const}-10)--(\ref{eq-const}-12) ensure that the turning of heading angle is penalized.  (\ref{eq-const}-13) ensures that the actions corresponding to higher Q-function values have a higher probability being executed.

We select the discount factor $\gamma=0.99$ for the MDP. This parameter determines the effective time horizon. A samll $\gamma$ would result in a greedy policy, which is undesirable. We solve this MDP using Gaussian Process Dynamic Programming \cite{deisenroth2009gaussian}.
\subsection{Prediction}
Once the model has been learned, we can infer the intention of the pedestrian from a partially observed trajectory $t_p$ using Eq. (\ref{intent_post}). Then we predict future trajectory
\begin{equation}
p(t_f|t_p,R)=\sum_{\lambda}p(t_f|\lambda,R)p(\lambda|t_p,R),
\label{generative}
\end{equation}
where $p(t_f|\lambda,R)$ could be represented as sample trajectories defined by the MDP. Eq. (\ref{generative}) is the generative process for the trajectories. In order to make prediction of future trajectories, we first sample the goal ${\lambda}_s$ from the discrete distribution over the intention $p(\lambda|t_p,R)$. Then we sample trajectories according to $p(t_f|{\lambda}_s,R)$, which is defined in Eq.~(\ref{log-ll}). Therefore, the distribution over future trajectory is a mixture distribution. 

\section{Results and Evaluations}
\subsection{Dataset}
\begin{figure}[t]
\centering
\includegraphics[width=0.5\textwidth]{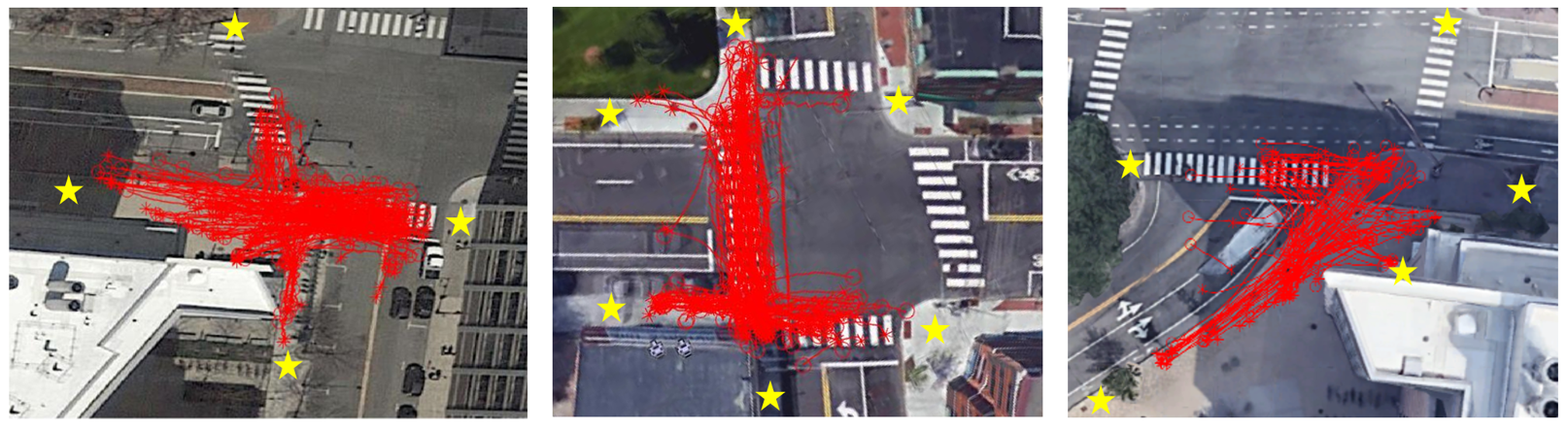} 
\caption{The intersection A (left) and the intersection B (right) for transfer capability evaluation, the yellow stars represent the manually selected goal locations}
\label{inter_st_view}
\end{figure}
We evaluated our algorithm at three intersections in Cambridge, Massachusetts, shown in Fig. \ref{inter_st_view}. Intersection A (on Ames Street) has 218 trajectories (186 training trajectories and 32 test trajectories, shown on the left). Intersection B (on Broadway) has 132 trajectories (100 training trajectories and 32 test trajectories, shown in the middle). Intersection C (on Vassar Street) has 134 trajectories (114 training trajectories and 20 test trajectories)

%The angles between the two crosswalks are different between these two intersections. As a result, algorithms that solely rely on discovering the pattern of the trajectories are not expected to work very well in the unseen intersection, which makes transferring difficult.

\begin{figure*}[t]
\centering
\includegraphics[width=0.65\textwidth]{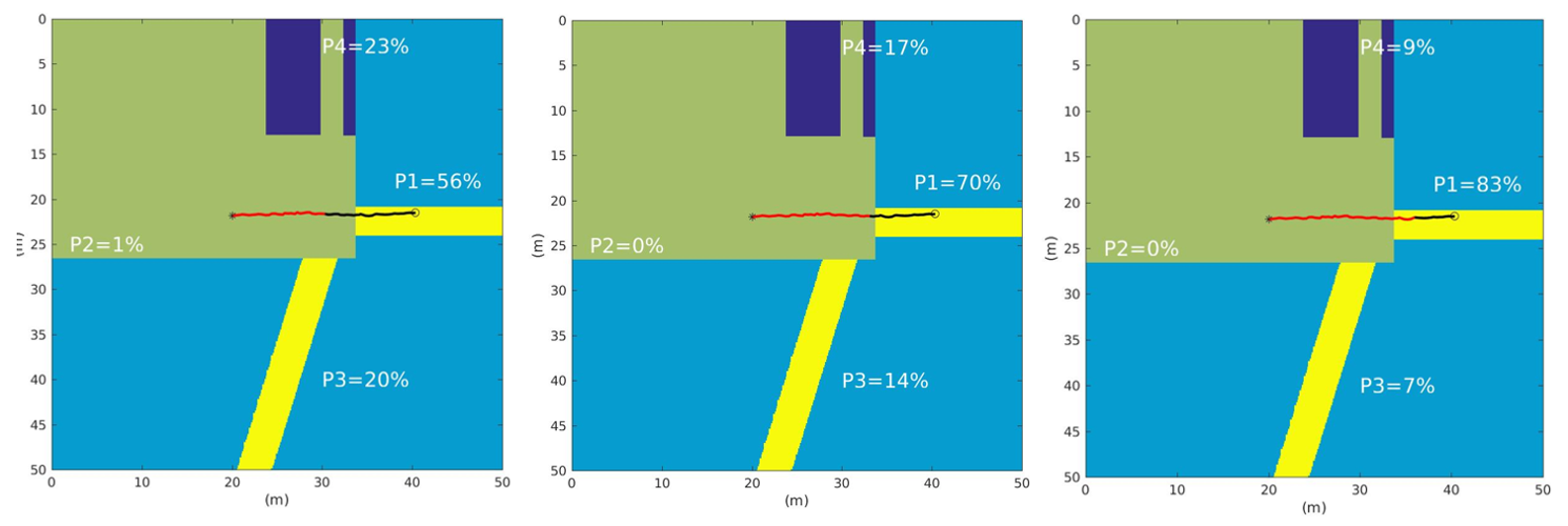}   
\caption{Intention recognition example: From left to right shows increased confidence of the prediction (shown as the probability corresponding to the four goals, P1, P2, P3, P4) as the length of the observed trajectory (the red segment) increases}
\label{tra3}
\includegraphics[width=0.65\textwidth]{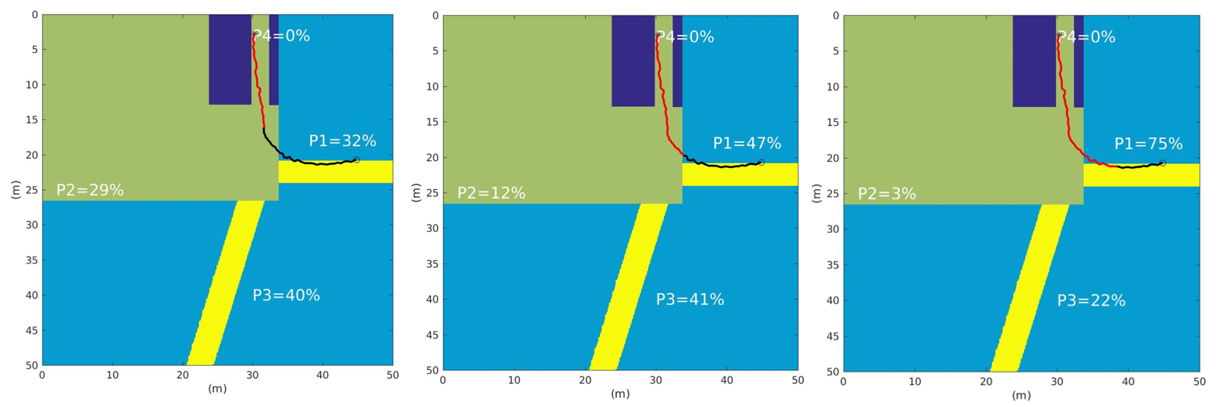}   
\caption{Intention recognition example: From left to right shows increased confidence of the prediction (shown as the probability corresponding to the four goals, P1, P2, P3, P4) as the length of the observed trajectory (the red segment) increases}
\label{tra11}
\end{figure*}

Figures \ref{tra3} and \ref{tra11} show two examples of intention recognition based on partial trajectory using the proposed algorithm. The algorithm infers the probability distribution over the four possible goal locations (denoted by P1, P2, P3, and P4) from the observed segment (red) of the trajectory at intersection A. As we increase the length of the red segment, the prediction accuracy increases.  

%\begin{figure}[h]
%\centering
%\includegraphics[width=0.75\columnwidth]{figure/training_test_data.eps}   
%\caption{Training (black) and test (red) trajectories: start from '*', end at 'o'}
%\label{trajectories}
%\includegraphics[width=\textwidth]{figure/tra11.png}   
%\caption{Illustration of the over-confidence issue in constant reward models}
%\label{tra11}
%\end{figure}

\subsection{Model Transfer}
To make prediction at an unseen intersection, both the model parameters and the goal locations must be specified. We assume that the model parameters $\theta=\{w,C_{\phi},\beta,\alpha,\eta\}$, learned from one intersection, are applicable to the unseen intersection, while the goal locations are specific to each intersections, which is not transferable. In order to make prediction at the new intersection, the goal locations are selected manually (see Fig. \ref{inter_st_view}) based on the semantic context. Once the model parameters and the goal locations are specified for the new intersection, the prediction process described in Section III. D is applied to predict the distribution of future trajectories. 

\subsection{Evaluation metric}
Given a partial trajectory, our algorithm first infers the intention of the pedestrian, which is a discrete probability distribution $p(\lambda|t_p)$ that specifies the probability of each potential goal location as the intention of the pedestrian. The MDP also implicitly defines a distribution over the possible trajectories given the intention and the reward function $p(t_f|\lambda,R)$. We define an accuracy metric EMHD (Expected Modified Hausdorff Distance) in Eq. (\ref{EMHD}) to evaluate the performance of our prediction. MHD \cite{dubuisson1994modified} is a distance defined over two sets of points with equal length and dimension. In this work, it is used to measure the similarity between the predicted trajectory and the true trajectory in the dataset. 
\begin{equation}
\begin{aligned}
EMHD&=\int MHD(t_f,t_{data})p(t_f|t_p,R)dt_f\\
& \approx \sum^N_{i=1} \frac{1}{N} MHD(t^i_f,t_{data}),
\end{aligned}
\label{EMHD}
\end{equation}
where $t_{data}$ is the trajectory present in the dataset. $t^i_f$, where $i=1,2,\ldots,N$, are samples of the future trajectories from the predictive model by first sampling the intention from $p(\lambda|t_p)$ and then sampling the trajectories from $p(t_f|\lambda,R)$, and $N$ is the total number of the samples used to approximate the integral. In this work, we choose $N=100$. We have verified that this choice ensures that the approximation error is under 1\%.
%\begin{table}[ht] 
%\caption{Accuracy on the test set} % title of Table 
%\centering      % used for centering table 
%\begin{tabular}{c c c c c}  % centered columns (4 columns) 
%\hline\hline                        %inserts double horizontal lines 
%Partial trajectory length &25\% &50\% &65\%&80\%\\ [0.3ex] % inserts table %heading 
%\hline                    % inserts single horizontal line 
%EMHD (m) & 2.76 & 1.67 &1.11 &0.54  \\   % inserting body of the table 
%\hline     %inserts single line 
%\end{tabular}
%\label{tEMHD}  % is used to refer this table in the text 
%\end{table}

%\indent Table \ref{tEMHD} shows the average EMHD over the test trajectories, predicted based on different lengths of the partial trajectories. As we increase the length of the partial trajectory, the prediction becomes easier. The algorithm is more confident of its intention inference, and the trajectory accuracy also increases, thus resulting in a lower EMHD. \\
\subsection{Comparison with baseline}

We used ASNSC (augmented semi-nonnegative sparse coding) \cite{chen2016augmented} as a baseline algorithm for comparison. ASNSC learns a part-based trajectory representation and creates GP velocity fields for motion prediction. ASNSC has been shown to outperform the previous GP-based clustering algorithm, DPGP \cite{joseph2011bayesian}.

We trained our algorithm at both intersections, and tested at both intersections. As the baseline algorithm ASNSC \cite{chen2016augmented} is not designed for transfer, in order to make a fair comparison, we trained ASNSC at all the three intersections. We then used the prediction accuracy of test data at the same training intersection as a baseline. For each test trajectory, we used the initial 2.5 s of the trajectory as the input to the algorithms, and compare the predictions of the next 5 s with the ground truth. Trajectories longer than 7.5 s are truncated. The prediction accuracy and computation time are shown in Table \ref{comparison_transferability}, \ref{comparison} and \ref{comparison_C}.

\begin{table}[t] 
\caption{Comparison between our algorithm and ASNSC at the training intersection A} % title of Table 
\centering      % used for centering table 
\begin{tabular}{c c c}  % centered columns (4 columns) 
\hline\hline                        %inserts double horizontal lines 
Algorithm &EMHD (m) &CPU time (s)\\ [0.3ex] % inserts table %heading 
\hline                    % inserts single horizontal line 
Our algorithm (trained on A) & 1.44 & 0.61   \\
Our algorithm (trained on B) & \bf{1.36} & 0.61   \\   % inserting body of the table
Our algorithm (trained on C) & 1.67 & 0.61   \\
ASNSC (trained on A) & 2.44 & \bf{0.04}   \\   % inserting body of the table
%CASNSC & 2.38 & 0.3  \\   % inserting body of the table
\hline     %inserts single line 
\end{tabular}
\label{comparison_transferability}  % is used to refer this table in the text 
%\end{table}
%
%\begin{table}[t] 
\caption{Comparison between our algorithm and ASNSC at the training intersection B} % title of Table 
\centering      % used for centering table 
\begin{tabular}{c c c}  % centered columns (4 columns) 
\hline\hline                        %inserts double horizontal lines 
Algorithm &EMHD (m) &CPU time (s)\\ [0.3ex] % inserts table %heading 
\hline                    % inserts single horizontal line 
Our algorithm (trained on A) & 0.87 & 0.60   \\
Our algorithm (trained on B) & \bf{0.84} & 0.60   \\   % inserting body of the table
Our algorithm (trained on C) & 1.07 & 0.60   \\
ASNSC (trained on B) & 1.03 & \bf{0.04}   \\   % inserting body of the table
%CASNSC & 2.38 & 0.3  \\   % inserting body of the table
\hline     %inserts single line 
\end{tabular}
\label{comparison}  % is used to refer this table in the text 

\caption{Comparison between our algorithm and ASNSC at the training intersection C} % title of Table 
\centering      % used for centering table 
\begin{tabular}{c c c}  % centered columns (4 columns) 
\hline\hline                        %inserts double horizontal lines 
Algorithm &EMHD (m) &CPU time (s)\\ [0.3ex] % inserts table %heading 
\hline                    % inserts single horizontal line 
Our algorithm (trained on A) & 1.70 & 0.35   \\
Our algorithm (trained on B) & \bf{1.67} & 0.35   \\   % inserting body of the table
Our algorithm (trained on C) & 2.03 & 0.35   \\
ASNSC (trained on C) & 3.60 & \bf{0.02}   \\   % inserting body of the table
%CASNSC & 2.38 & 0.3  \\   % inserting body of the table
\hline     %inserts single line 
\end{tabular}
\label{comparison_C}  % is used to refer this table in the text 

\end{table}

We see that our algorithm achieves higher accuracy than ASNSC does in general. We think that the reasons that our algorithm outperforms ASNSC are: (1) We use semantic features (road, obstacle, crosswalk, and sidewalk); (2) We explicitly specify several potential goal locations and use radial basis reward function to direct the trajectories towards the goals, while this is not done in ASNSC.

The most accurate prediction at each of the three intersections is made by the model trained at intersection B. The model trained at intersection A makes a less accurate prediction, and the model trained at intersection C is the least accurate, but still outperforms ASNSC.  One possible explanation is that the training data at intersection B is better for our algorithm to learn the model parameter than the other two intersections, as the trajectories at intersection B follows the semantic contexts more strictly.
 %\todo{challenge for which algorithm?}

\balance
\section{Conclusion}
In this paper, we investigated transfer learning for pedestrian motion prediction at intersections. We proposed a rule of feature selection for the design of transferable motion prediction algorithms. We designed an IRL-based algorithm based on this rule and demonstrated its transferring capability. In our algorithm, we incorporate knowledge about the pedestrian motion: (1) Pedestrian motions are directed by the goal locations; (2) Pedestrians have different preferences over different semantic contexts; (3) Pedestrians tend to keep their heading angles. This structural knowledge is not included in ASNSC, while it needs to discover this knowledge from the data, which could be challenging for ASNSC. We showed that the proposed algorithm achieves good prediction accuracy in both non-transferring task and transferring task, compared with ASNSC, at the cost of increased computation time.

One limitation of our method is that the goal locations should be selected based on the semantic context. We did this selection manually in this paper, and we have not provided a principled way for this selection. For future work, we would like to investigate the sensitivity of the prediction result on the goal locations.

%\todo{please compare with previous algorithms appropriately, also, it needs more metric to compare}

\section*{Acknowledgment}
This work is supported by Ford Motor Company. The authors want to thank Nikita Jaipuria and Michael Everett for the data collection and discussion about the results in this paper.

%This approach learns a mapping from the semantic features to the reward function as well as a pedestrian model that describes the turning preference. It is assumed that this mapping and pedestrian model can be generalized to the unseen intersections. 

\bibliographystyle{IEEEtran} 
\bibliography{pedestrian}
\end{document}